# Physics-Informed Machine Learning of Argon Gas-Driven Melt Pool Dynamics

R. Sharma[a,b], W. Grace Guo[b,c], M. Raissi[d], Y.B. Guo[a,b]

[a] Dept. of Mechanical and Aerospace Engineering, Rutgers University-New Brunswick, Piscataway, NJ 08854, USA
[b] New Jersey Advanced Manufacturing Institute, Rutgers University-New Brunswick, Piscataway, NJ 08854, USA
[c] Dept. of Industrial and Systems Engineering, Rutgers University-New Brunswick, Piscataway, NJ 08854, USA
[d] Department of Applied Mathematics, University of Colorado, Boulder, CO 80309, USA

**Abstract**

Melt pool dynamics in metal additive manufacturing (AM) is critical to process stability, microstructure formation, and final properties of the printed materials. Physics-based simulation including computational fluid dynamics (CFD) is the dominant approach to predict melt pool dynamics. However, the physics-based simulation approaches suffer from the inherent issue of very high computational cost. This paper provides a physics-informed machine learning (PIML) method by integrating neural networks with the governing physical laws to predict the melt pool dynamics such as temperature, velocity, and pressure without using any training data on velocity. This approach avoids solving the highly non-linear Navier-Stokes equation numerically, which significantly reduces the computational cost. The difficult-to-determine model constants of the governing equations of the melt pool can also be inferred through data-driven discovery. In addition, the physics-informed neural network (PINN) architecture has been optimized for efficient model training. The data-efficient PINN model is attributed to the soft penalty by incorporating governing partial differential equations (PDEs), initial conditions, and boundary conditions in the PINN model.

## 1. Introduction

Metal additive manufacturing (AM) is a very important manufacturing technique that has the potential to revolutionize the mechanical and aerospace industries owing to its capability of printing complex shapes by using the digital model. However, despite its potential, metal AM has not yet reached its expected level of usage in industries, in part due to a lack of accurate prediction of the properties of printed components. For example, in laser powder bed fusion (LPBF), the layer of metal powder is scanned by a laser heat source which converts the metal powder to liquid, which eventually solidifies and converts to the final product. Accurate thermal history prediction is crucial for LPBF, as all other phenomena, including thermal residual stress and microstructure, depend on it. The melt pool dynamics play a very important role in the development of the thermal map for LPBF. Many factors influence the melt pool dynamics in LPBF such as the unique thermal cycle of rapid heating and solidification, steep temperature gradient and high cooling rate, evaporation, surface tension, natural convection, Marangoni convection, vapor recoil pressure, and Argon flow over the melt pool. Several researchers have developed computational models to better understand melt pool dynamics, incorporating these complex phenomena [1-5].

Physics-based simulation such as computational fluid dynamics (CFD) is the key method to model melt pool dynamics (Figure 1). Li et al. [6] utilized a 2D model to examine the melting and solidification. The model incorporates the Volume of Fluid (VOF) method to identify the interface during the process. Although the model considers the impact of surface tension, it does not include the Marangoni and recoil forces. Gurtler et al. [3] developed a 3D transient model to simulate the dynamics of the melt pool using the OpenFOAM. The model is simplified as it excludes the impact of surface tension, Marangoni, and recoil forces. Still, the model was able to accurately reproduce the essential features of the laser beam melting process. Additionally, Tseng and Li [4] investigated the influence of surface tension, Marangoni, and recoil forces on the shape of the interface during selective laser melting (SLM). De Baere et al. [2] developed a numerical model for the SLM process, which incorporates both thermo-fluid and metallurgical elements. This model can determine the solidification parameters, such as the temperature gradient, cooling rate, and growth velocity, and classify the grain structure as either columnar or equiaxed. The predicted morphology is then fed into a microstructural model, which predicts the final microstructure after heat treatment is completed.

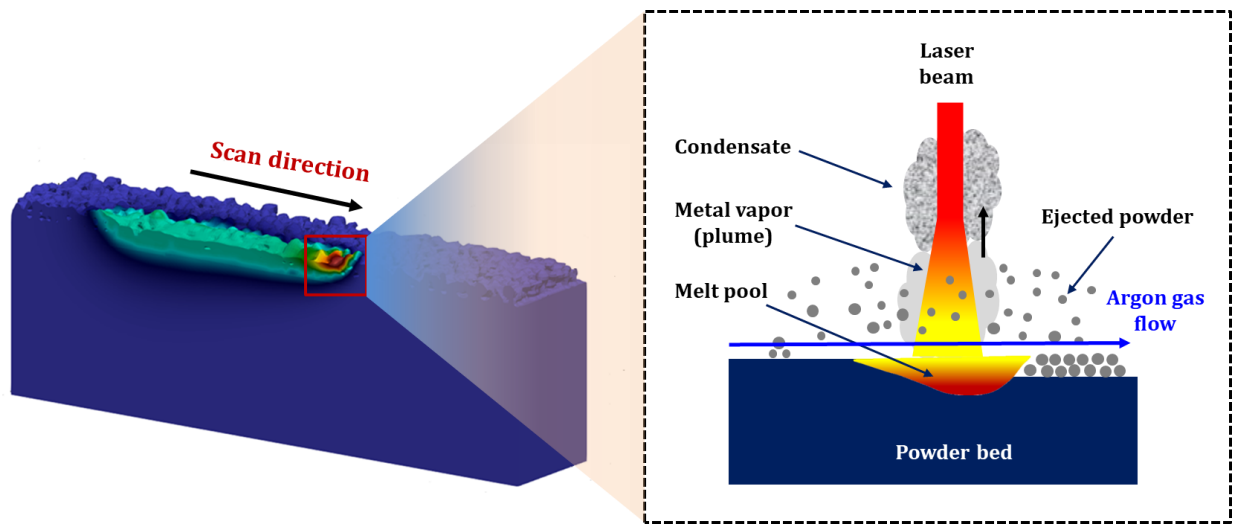

**Fig. 1:** Schematic of Argon flow over the melt pool

The presence of a high-temperature gradient in the melt pool results in a plume of fine powder particles resulting from the evaporation of the material [7]. The flow of inert gas over the melt pool has a significant influence on the LPBF process as it not only prevents oxidation of the metal but also helps in the removal of this microscale plume [8-11]. Also, at the same time, the velocity of inert gas should be optimum so that the plume will get away and there is no pickup of powder particles from the spread powder layer [12]. Anwar and Pham [13] observed that the mechanical properties like tensile strength of the material will increase when the material is scanned against the gas flow, and the speed of inert gas increases from 30% to 60% of the pump setting. Apart from the speed of the gas flow, uniformity of the gas flow is very important for the LPBF process. Researchers developed several computational fluid dynamics (CFD) models to enhance the uniformity of the gas flow [2-7]. Ferrar et al. [14] used the CFD software suite Fluent to improve the uniformity of gas flow in a commercial LPBF system, which resulted in enhanced performance. The improvement in gas flow uniformity was achieved by reducing the porosity and increasing the strength of the resulting material. Alquaity and Yilbas [15] developed a numerical model to study the interaction between gas flow and spattered particles. It was observed that the bigger powder particles reach beyond the build region only at specific gas ejection angles and higher gas flow rates. Furthermore, this inert gas flow has a significant impact on the fluid velocity of the top layer of the melt pool, which as a result drives the fluid flow in the whole melt pool. This will result in a change in the width and depth of the melt pool. Though this is an important phenomenon, it has not been explored in-depth in the literature and is one focus of this study.

One major challenge associated with the above physics-based modeling is very high computational cost as LPBF is a multiphysics problem that governs by the coupled energy equation and the highly non-linear Navier Stokes equation along with source and sink terms. Data-driven approaches are becoming attractive for solving multiphysics problems in biomedical [16, 17], language processing [18, 19], and computer vision [20, 21]. Many machine learning (ML) algorithms have been used in the past to understand fluid mechanics [22-25]. Some LPBF printers are equipped with high-speed cameras and pyrometers which can be used to collect data and train the ML models. Recently, some researchers studied melt pool characteristics through machine learning models[26-34]. However, the present ML algorithms that rely solely on data-driven approaches have limitations in terms of their lack of interpretability, computational and storage requirements, and dependence on large training data sets for achieving high prediction accuracy. Specifically, they suffer from a "black box" nature due to the absence of process physics and explainability and are inherently time-consuming operations [35, 36]. Thus, there is a need for more advanced approaches that can address these challenges by leveraging the physical laws underlying the melt pool dynamics to improve the overall performance and efficiency of ML models.

Physics-informed neural networks (PINNs) are an emerging technique that utilizes the governing equations of the process dynamics and results in fast training with small training data[37]. Initially, vanilla feed-forward neural networks were used for PINNs, but later researchers explored multiple feed-forward networks[38, 39], convolution neural networks [40, 41], recurrent neural networks [42, 43], and Bayesian neural networks[44]. PINNs have been used to solve standard fluid mechanics problems[45-47]. Recently, the PINN model has been developed for LPBF, but it needs high-resolution training data for temperature and velocity fields to predict the velocity field [48, 49]. It leads to the same challenge of solving the highly non-linear Navier Stokes (NS) equation through physics-based simulation to generate massive training data.

In this study, a PINN model is developed to predict the 3D melt pool dynamics due to Argon gas-driven shear flow on the top of the melt pool without solving the non-linear Navier Stokes equations. The unique scientific merits of the PINN model over the already available model are: 1) it can be trained without using training data on fluid velocity that avoids the solution of the nonlinear Navier-Stokes equation which results in a computationally efficient model. 2) PINNs can be trained in much less time compared to conventional ML models. 3) Compared to conventional ML models, PINNs need less training data because it gives an extra penalty to the model by incorporating governing PDEs, initial conditions, and boundary conditions. 4) PINNs are capable of inferring difficult-to-determine parameters (e.g., Reynolds number and Peclet number) of the governing equations through the data-driven discovery approach. The paper is divided into four sections. First, the governing equations for the multiphysics problem of Argon gas-driven shear flow are presented. Second, the results of the forward prediction are discussed, where temperature data and equation parameters, such as Reynolds and Peclet numbers, are provided to the PINN model to predict the melt pool velocity field. Third, the results of the inverse learning are presented, where both temperature and velocity data are provided to the PINN model to infer the governing equation parameters. Lastly, a general method to solve new multiphysics problems is discussed.

## 2. Coupled Forward-Inverse Learning Framework

Figure 2 illustrates the research framework which encompasses the forward and inverse problems. For the forward problem, a neural network utilizes temperature data from collocation points to predict velocity and pressure values at the corresponding points. In the inverse problem, the neural network utilizes the temperature and velocity data from collocation points to estimate equation parameters. This framework enables efficient analysis and prediction of complex phenomena by integrating neural networks for both forward and inverse problem-solving. In the subsequent subsections, the governing equations for the process and an overview of the fundamental principles behind PINNs are introduced.

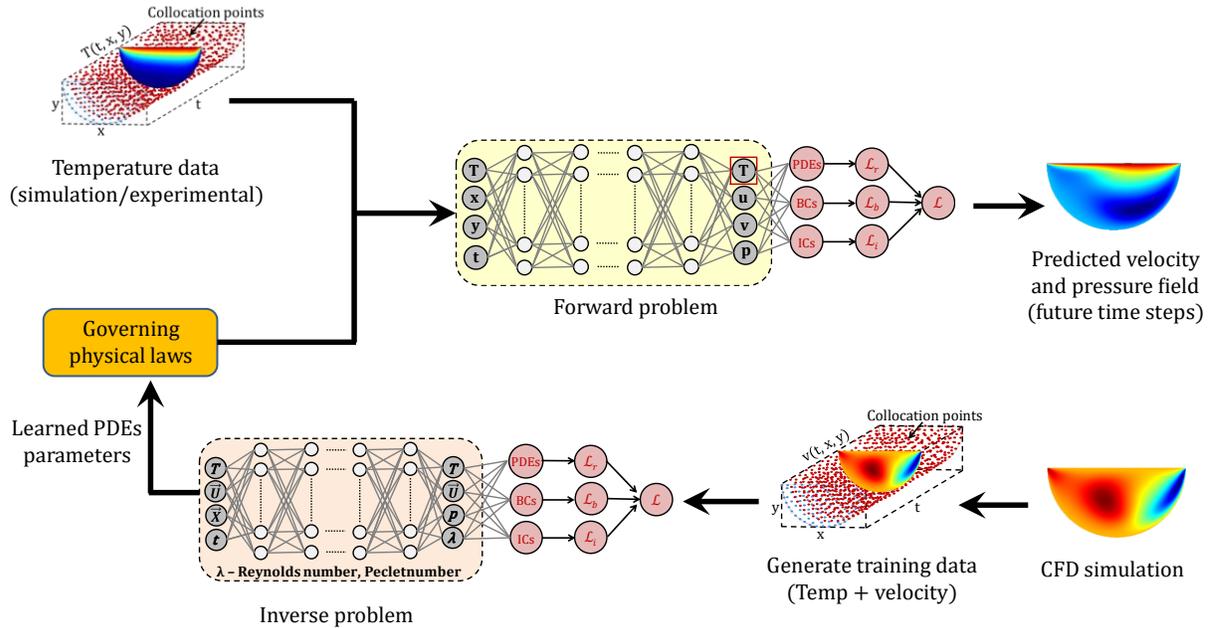

**Fig. 2:** Coupled forward-inverse learning framework.

*2.1 Governing equations*

Figure 3 illustrates the Argon flow over the top surface of the melt pool in LPBF. The focus of this study is to explore the capability of PINN to predict the fluid flow in the melt pool without numerically solving the non-linear Navier Stokes equations. Therefore, some reasonable assumptions are made to carry out this study. The shape of the melt pool is assumed to be a hemisphere with the ratio of diameter to depth as 2:1 as this number is approximately observed by Agarwal et. al [50]. Further, to simplify the analysis, a single cross-section is analyzed in this study as important melt pool dynamics can be observed from the cross-section view (Figure 2). The effect of Argon flow on the melt pool dynamics and temperature distribution is considered. COMSOL Multiphysics software is used to generate the high-resolution dataset which is required to train the PINN model. The top boundary is assumed to be moving of Ar flow with a velocity of 5 m/s [7]. The curved boundary has a no-slip condition. In LPBF, the laser heat flux is interacting on the top boundary, therefore, the top boundary is assumed to have the highest temperature. The curved boundary of the melt pool is assumed to have the lowest temperature of the melt pool. In this study, the highest temperature is 3090 K, which is the evaporation temperature of SS 316L powders, and the lowest temperature on the curved boundary is 1723 K, which is the liquidus temperature of SS 316L. With the above assumptions, the melt pool dynamics can be defined by the following coupled PDEs:

$$\frac{\partial T^*}{\partial t^*} + u^* \frac{\partial T^*}{\partial x^*} + v^* \frac{\partial T^*}{\partial y^*} - \frac{1}{Pe} \nabla^2 T^* = 0 \quad (1)$$

$$\frac{\partial u^*}{\partial t^*} + u^* \frac{\partial u^*}{\partial x^*} + v^* \frac{\partial u^*}{\partial y^*} + \frac{\partial P^*}{\partial x^*} - \frac{1}{Re} \nabla^2 u^* = 0 \quad (2)$$

$$\frac{\partial v^*}{\partial t^*} + u^* \frac{\partial v^*}{\partial x^*} + v^* \frac{\partial v^*}{\partial y^*} + \frac{\partial P^*}{\partial y^*} - \frac{1}{Re} \nabla^2 v^* = 0 \quad (3)$$

$$\frac{\partial u^*}{\partial x^*} + \frac{\partial v^*}{\partial y^*} = 0 \quad (4)$$

where $\quad Re = \frac{\rho D U_{Ar}}{\mu}, \quad Pe = \frac{U_{Ar} D \rho c_p}{K} \quad (5)$

$T^*$, $u^*$, $v^*$, and $p^*$ represent the normalized melt pool temperature, x-component of velocity, y-component of velocity, and pressure, respectively. $Pe$ and $Re$ represent the Peclet number and Reynolds number and $\nabla^2$ is the Laplace operator. With the value of the velocity at the top boundary ($U_{Ar}$), the diameter of the melt pool ($D$), maximum and minimum values of temperature, and the material properties of SS 316L including density ($\rho$), viscosity ($\mu$), heat capacity ($C_p$), and thermal conductivity ($K$) from Table 1, the values of Peclet and Reynolds number can be calculated as 77 and 541 respectively.

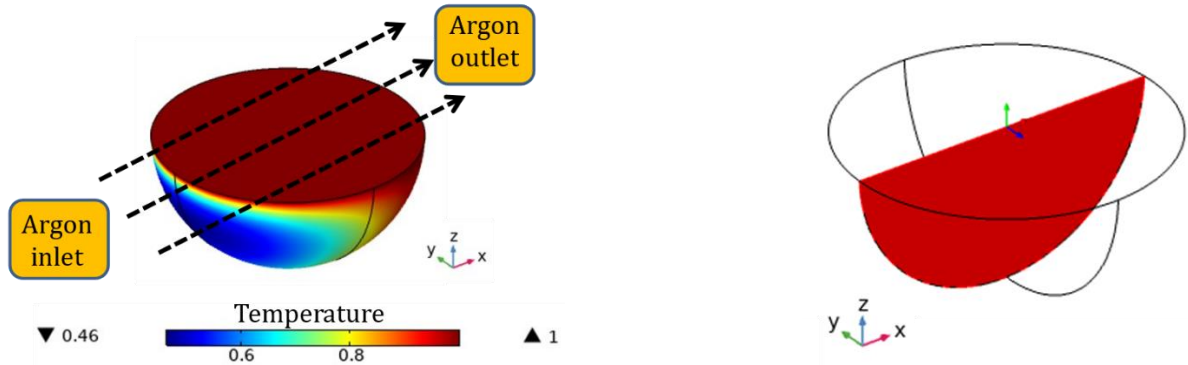

**Fig. 3: (a)** actual 3D melt pool **(b)** cross-section analysis.

**Table 1:** Material properties of SS 316L [51].

| | |
|---|---|
| Density of liquid metal (Kg/m³) | 6500 |
| Viscosity (m²/s) | 0.006 |
| Heat capacity of liquid metal (Kg-m/s³) | 830 |
| Thermal conductivity of liquid metal (Kg-m/s³K) | 35 |

Page **5** of **18**

The simulation has been performed for 0-13 seconds and the steady-state flow is achieved at the last time step (13th second). In this study, the PINN model is trained using data at the unsteady state (0-13 seconds) data and the predictive power is demonstrated by comparing the predicted solution with the ground truth at the steady state (13th second).

*2.2 PINN for multiphysics melt pool dynamics*

Figure 4 shows the architecture of the fully connected PINN used in this study. A neural network is a technique of computation that is inspired by the architecture of the human brain. In this study, a fully connected PINN is developed with 10 hidden layers and 50 neurons per hidden layer, and a learning rate of $10^{-3}$. These hyperparameters were selected from the literature [52]. This PINN framework utilizes the temperature field and physical laws in the form of governing equations to predict the velocity field of the melt pool without any training data. For a fully connected neural network, the output of the current ($n^{th}$) layer is related to the previous $(n-1)^{th}$ layer by the following relation:

$$z_n = \sigma_n(w_n^T z_{n-1} + b_n) \tag{6}$$

where $w_n$ and $b_n$ are the weights and biases for the current layer and $\sigma_n$ is the swish activation function which is given by:

$$\sigma(x) = swish(x) = x\, sigmoid(x) = x/(1 - e^{-x}) \tag{7}$$

The PINN takes $\{t^n, x^n, y^n, T^n\}_{n=1}^{N}$ as input which represents the spatiotemporal coordinates of the collocated points in the computational domain and the temperature data on these points. In this study, temperature data at 13,500 scattered points per time step is provided to PINN for 13 different time points between 0 to 13 seconds to train the model. The output of the neural network is $\{T^n, u^n, v^n, p^n\}_{n=1}^{N}$ which are the predicted values of temperature, u-velocity, v-velocity, and pressure respectively at the same collocated points.

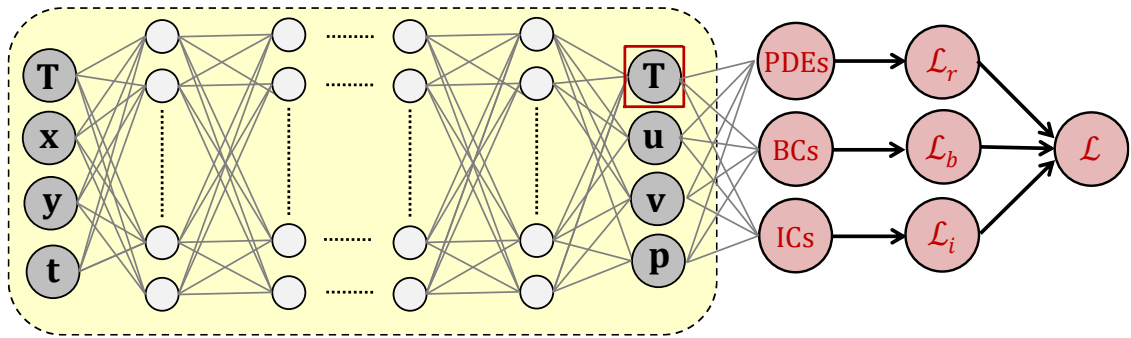

**Fig. 4:** PINN architecture with PDEs, boundary conditions and initial conditions loss.

It is worth noting that only temperature data at collocated points, initial conditions, and boundary conditions are used to predict the velocity field of the whole melt pool. This predictive capability is enabled by the PINN architecture which allows the prediction of temperatures and velocity field using the data loss. The data loss is typically measured by the mean squared error loss for regression over the temperature field:

$$\mathcal{L}_{Data} = \frac{1}{N}\sum_{n=1}^{N}|T_{pred}^{n}(t,x,y) - T_{exact}^{n}(t,x,y))|^2 \qquad (8)$$

where $T_{pred}^{n}$ and $T_{exact}^{n}$ are the predicted and true temperature values at the collocation points respectively and N represents the sampling points for temperature.

Since solving the forward problem of learning the velocity field requires initial and boundary conditions. The initial conditions are enforced with the following loss functions:

$$\mathcal{L}_{u,IC} = \frac{1}{Q}\sum_{q=1}^{Q}|u_{IC,pred}^{q}(t^i,x^i,y^i) - u_{IC,exact}^{q}(t^i,x^i,y^i)|^2 \qquad (9)$$

$$\mathcal{L}_{v,IC} = \frac{1}{Q}\sum_{q=1}^{Q}|v_{IC,pred}^{q}(t^i,x^i,y^i) - v_{IC,exact}^{q}(t^i,x^i,y^i)|^2 \qquad (10)$$

where $(t^i, x^i, y^i)$ are the spatiotemporal data at the initial step, $U_{IC,pred}^{q}$ and $U_{IC,exact}^{q}$ are the component of the velocity field at the first time step and $Q$ represents the sampling points for the velocity field at the initial time step. To enforce the boundary conditions at the wall, the following equations are used:

$$\mathcal{L}_{u,top} = \frac{1}{P}\sum_{p=1}^{P}|u_{BC,pred}^{p}(t^t,x^t,y^t) - u_{BC,exact}^{p}(t^t,x^t,y^t)|^2 \qquad (11)$$

$$\mathcal{L}_{v,top} = \frac{1}{P}\sum_{p=1}^{P}|v_{BC,pred}^{p}(t^t,x^t,y^t) - v_{BC,exact}^{p}(t^t,x^t,y^t)|^2 \qquad (12)$$

$$\mathcal{L}_{u,bottom} = \frac{1}{P}\sum_{p=1}^{P}|u_{BC,pred}^{p}(t^b,x^b,y^b) - u_{BC,exact}^{p}(t^b,x^b,y^b)|^2 \qquad (13)$$

$$\mathcal{L}_{v,bottom} = \frac{1}{P}\sum_{p=1}^{P}|v_{BC,pred}^{p}(t^b,x^b,y^b) - v_{BC,exact}^{p}(t^b,x^b,y^b)|^2 \qquad (14)$$

where $(t^t, x^t, y^t)$, and $(t^b, x^b, y^b)$ are the spatiotemporal point cloud on the flat top and curved bottom boundary of the domain respectively and $P$ represents the sampling points for the velocity field at the boundaries.

The unique feature of PINN is to utilize the governing equations of the melt pool dynamics. This can be done by defining the consistency loss. At the start of the training, the prediction of temperature and velocity is not accurate. The predicted temperature and velocity fields are used in the governing equations to calculate the residual given by:

$$f^T(t,x,y) = \frac{\partial T^*}{\partial t^*} + u^*\frac{\partial T^*}{\partial x^*} + v^*\frac{\partial T^*}{\partial y^*} - \frac{1}{Pe}\nabla^2 T^* \qquad (15)$$

$$f^u(t,x,y) = \frac{\partial u^*}{\partial t^*} + u^*\frac{\partial u^*}{\partial x^*} + v^*\frac{\partial u^*}{\partial y^*} + \frac{\partial P^*}{\partial x^*} - \frac{1}{Re}\nabla^2 u^* \qquad (16)$$

$$f^v(t,x,y) = \frac{\partial v^*}{\partial t^*} + u^*\frac{\partial v^*}{\partial x^*} + v^*\frac{\partial v^*}{\partial y^*} + \frac{\partial P^*}{\partial y^*} - \frac{1}{Re}\nabla^2 v^* \qquad (17)$$

$$f^m(t,x,y) = \frac{\partial u^*}{\partial x^*} + \frac{\partial v^*}{\partial y^*} \qquad (18)$$

where $f^T, f^u, f^v$, and $f^m$ corresponds to the residuals of energy, x-momentum, y-momentum, and mass conservation equations. When the predicted value is accurate then the residual becomes zero. The residual loss is given by:

$$\mathcal{L}_R = \frac{1}{M}\left\{\sum |f^T(t^c,x^c,y^c)|^2 + \sum |f^u(t^c,x^c,y^c)|^2 \\ + \sum |f^v(t^c,x^c,y^c)|^2 + \sum |f^m(t^c,x^c,y^c)|^2\right\} \qquad (19)$$

where $(t^c, x^c, y^c)$ are the collocation points where the temperature and velocity field data are observed. Finally, the total loss can be defined as:

$$\mathcal{L} = \underbrace{\mathcal{L}_{Data}}_{\text{Data loss}} + \underbrace{\mathcal{L}_{u,IC} + \mathcal{L}_{v,IC}}_{\text{IC loss}} + \underbrace{\mathcal{L}_{u,top} + \mathcal{L}_{v,top} + \mathcal{L}_{u,bottom} + \mathcal{L}_{v,bottom}}_{\text{BC loss}} + \underbrace{\mathcal{L}_R}_{\text{PDE residual}} \qquad (20)$$

$\mathcal{L}_{Data}$ helps to learn the model from the labeled data while other terms enable the training process fast by giving the extra penalty to the model if predicted values of parameters do not satisfy the initial conditions, boundary conditions, and governing equations. Since energy and momentum equations are coupled and temperature is a passive scalar whose distribution depends on the velocity field and not vice versa, therefore, it is possible to predict the velocity field by using only the temperature data on collocation points. To run this PINN model, NVIDIA A6000 GPU is used where 1.2 seconds are taken to perform 10 iterations. TensorFlow open library is used to calculate the derivatives of the process variables by the automatic differentiation (AD) technique. This technique is different and superior as compared to numerical differentiation. AD uses the chain rule to calculate the derivative which is accurate up to the machine's precision.

Finally, to measure the accuracy of the predictions, a relative $L_2$ error is calculated between the predicted quantity $p$ and the exact function $f$ as given by:

$$\in (p,f) = \left(\frac{1}{N}\sum_{i=1}^{N}[p(x_i) - f(x_i)]^2\right) / \left(\frac{1}{N}\sum_{i=1}^{N}[f(x_i) - \frac{1}{N}\sum_{i=1}^{N}f(x_i)]^2\right) \qquad (21)$$

where $\{x_i = 1, 2,\ldots, N\}$ are the collocation points scattered in the whole domain.

## 3. Result Analysis

To demonstrate the effectiveness of the proposed PINN algorithm, the effect of Argon flow on the melt pool dynamics is analyzed. For the forward prediction, the temperature data at different spatiotemporal coordinates are fed to the PINN, which in return predicts the temperature and velocity field on the same spatiotemporal coordinates. The model undergoes 24,000 iterations to get trained which approximately took 48 minutes. A batch size of 8,192 was used to perform this study. The results are compared at the last time step where the steady state is achieved. Firstly, the predicted results for the forward problem, convergence plots, the effect of boundary conditions, and the impact of training data resolution are discussed. Then, the capability of PINN to learn the equation parameters (Reynolds and Peclet number) is analyzed.

*3.1 Forward prediction of melt pool dynamics*

Figures 5, 6, and 7 show the comparison of predicted values of temperature, u-velocity, v-velocity, and pressure with the ground truth at the last time step respectively. The predicted results of temperature, u-velocity, and v-velocity match very well with the CFD results. Regarding the predicted pressure, it is worth noting that the qualitative prediction of pressure is very good though no training data on pressure is provided (Figure 7). However, the predicted pressure is shifted from the CFD simulated pressure by some constant value (it is 1 in this case). The difference in the magnitude of the predicted and exact pressure can be explained through the nature of the governing Navier-Stokes equation which contains only the derivative of the pressure due to which pressure prediction is always shifted from the exact value which is also observed in literature [53]. Overall, it shows the capability of the PINN model to predict the exact velocity and pressure fields without having any training data on it. PINN can become a very useful tool to calculate the velocity field for complex and computationally expensive multiphysics problems without solving the highly non-linear Navier-Stokes equation. Furthermore, in the future, this method can predict the melt pool dynamics in an LPBF process when the laser heat source is heating the powder and many complex physical phenomena such as melting, solidification, evaporation, natural and Marangoni convection, convective and radiative heat losses are involved. In that case, it can predict the melt pool dynamics by using only the scattered temperature data and without solving the non-linear Navier-Stokes equations as in physics-based simulation.

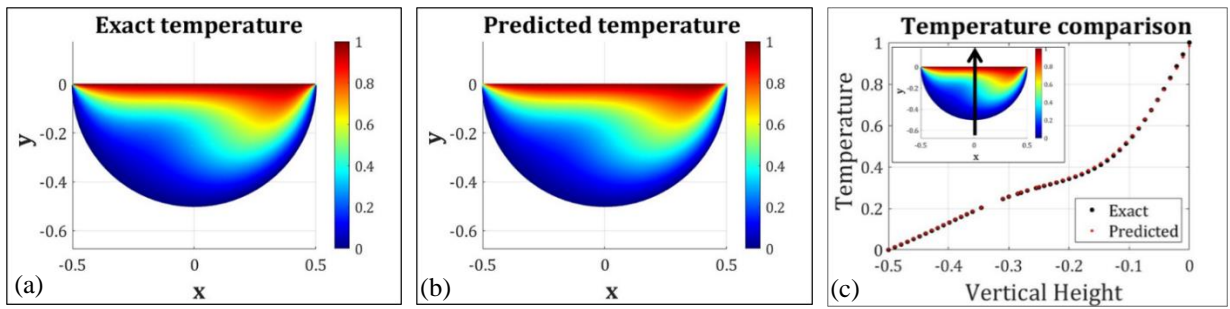

**Fig. 5:** Comparison of predicted temperature with the exact values at time =13s: (a) exact; (b) predicted; (c) exact vs predicted value of T, at the centre vertical line.

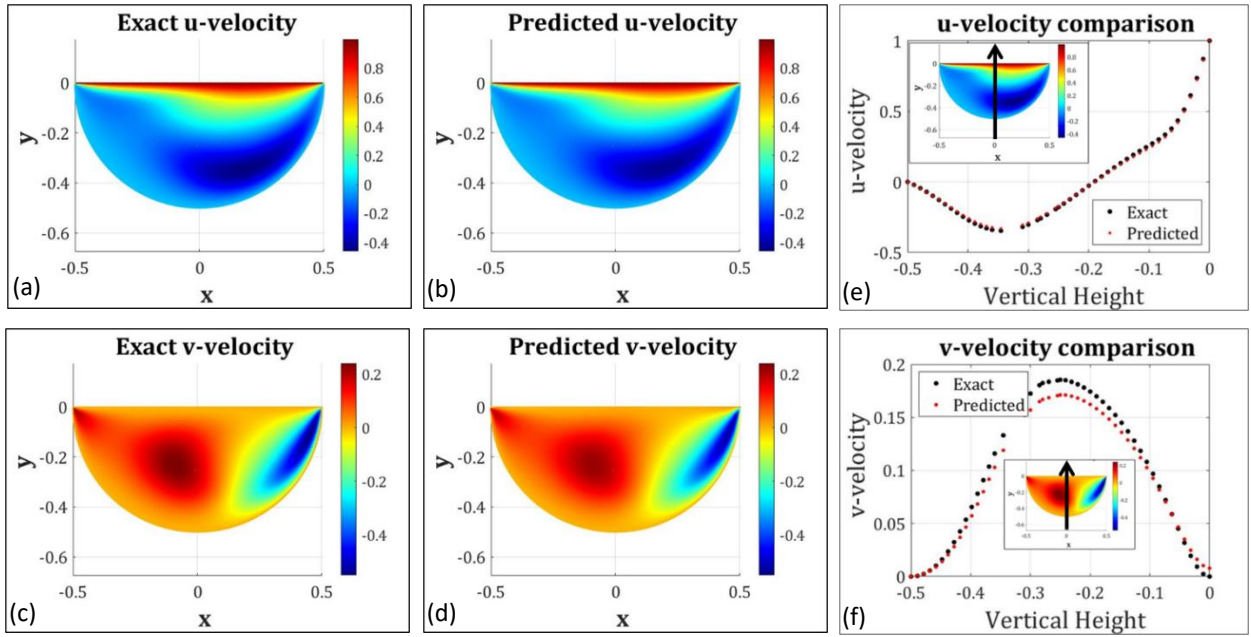

**Fig. 6:** Comparison of predicted results with the exact values u and v contours (a-d) at time =13s vs. the predicted value of u, and v (e-f) at the centre vertical line.

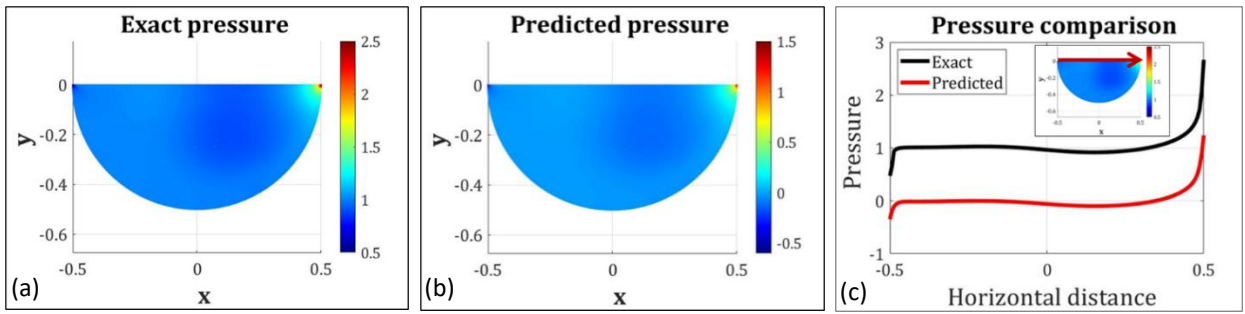

**Fig. 7:** Comparison of predicted temperature with the exact values at time =13s: (a) exact; (b) predicted; and (c) exact vs predicted value of T, at the horizontal line on the top.

Figure 8 shows the history of losses and L$_2$ errors for the training parameters. L$_2$ errors are calculated by using Eq. 11. These plots help to determine the total training time for the network. In this study, the PINN model gets sufficient training after 24,000 iterations. This is comparatively very efficient when compared to the conventional ML models, which is attributed to the extra penalty given to the PINN model in the form of loss functions addressing the residual of the governing equations, initial conditions, and boundary conditions.

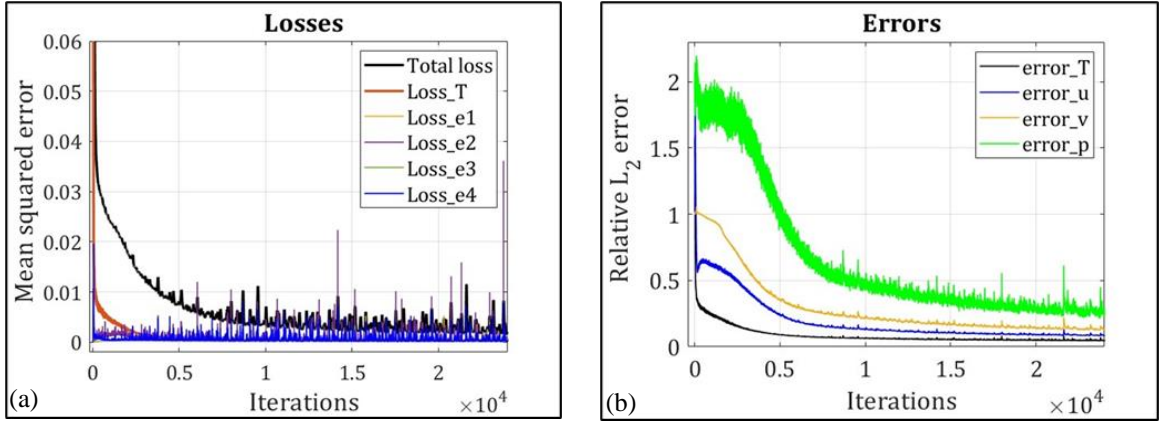

**Fig. 8:** History of losses (a) and $L_2$ errors (b).

### 3.1.1 Effect of boundary conditions on PINN prediction

For some simple problems, PINNs can predict fluid flow in the domain without using boundary conditions [52]. Figure 9 shows the prediction of temperature, velocity, and pressure field by the current PINN model without providing the velocity data on the boundaries for the current problem. The predicted temperature field still matches with the ground truth as temperature training data is provided to the model. But the velocity and pressure field are not matching with the ground truth. The inaccuracy in the prediction might be due to the presence of the steep gradients at the corners of the boundaries due to which the algorithm is not converging to a single solution. Therefore, it is a must to enforce boundary conditions during the training of the current model to get a single accurate solution.

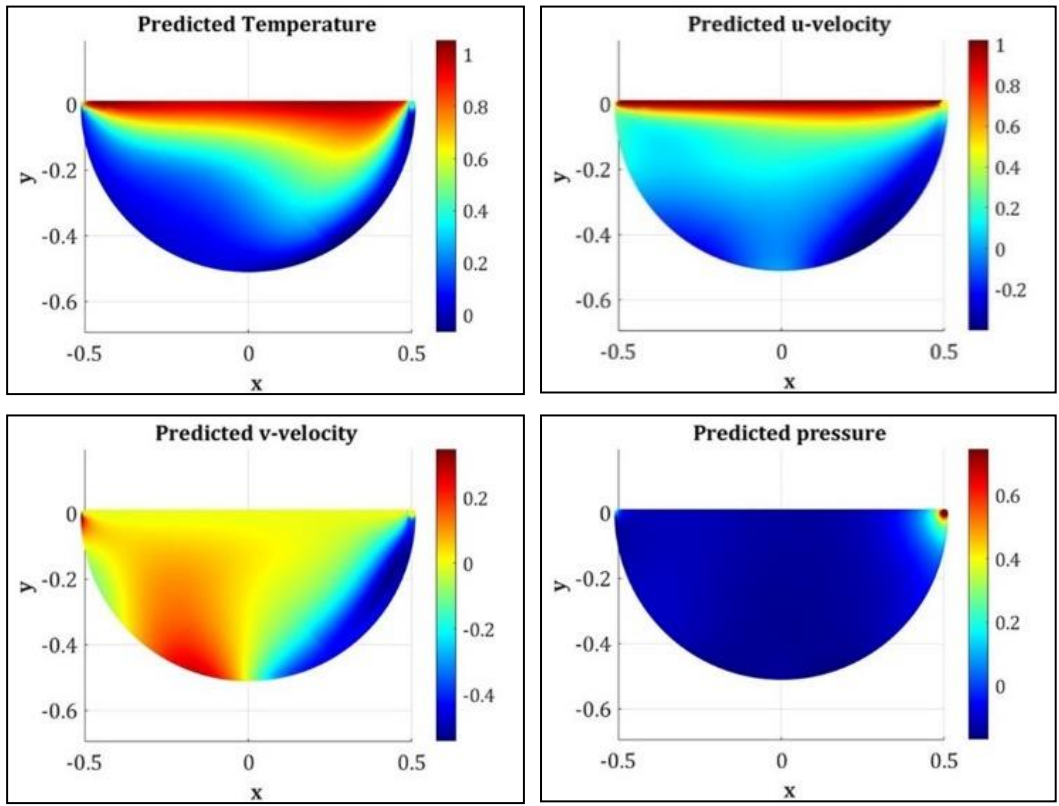

**Fig. 9:** Predicted results of T, u, v, and p contour at t = 13s without incorporating boundary condition.

### 3.1.2 Effect of training data on PINN prediction

It is mentioned that the PINNs are data and time-efficient models compared to conventional ML models in the above sections. To check the data efficiency of the PINN model, training is performed at different spatiotemporal resolutions. As shown in Figure 10, the PINN model is trained between 50-13,500 data points and 3-13 time points. It is observed that the PINN model can predict accurately up to 850 data and 7 time points. This is comparatively very data efficient as compared to the massive data required for a conventional pure data-driven ML model. This is due to the incorporation of governing PDEs, initial conditions, and boundary conditions in the PINN model, which provides an extra penalty to the model if the predicted solution does not satisfy the governing equations after each iteration of training.

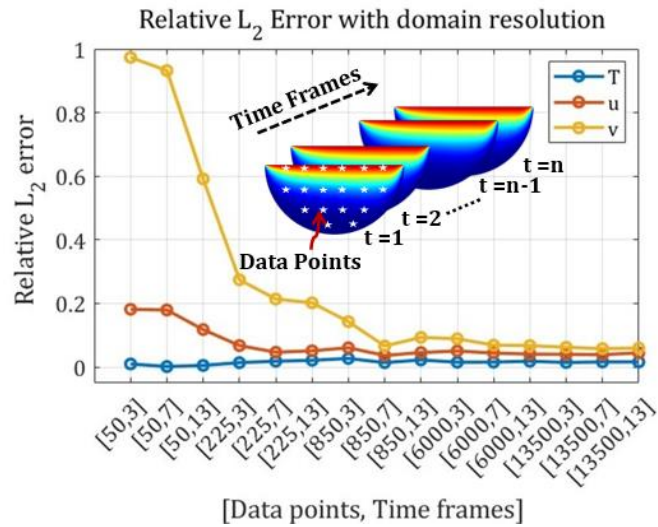

**Fig. 10:** Effect of spatial-temporal resolution on the relative $L_2$ error.

### 3.2 Inverse learning of model parameters

In many complex engineering problems, the governing equation parameters are unknown. For example, the Reynolds number and Peclet number in the non-linear Navier-Stokes equations are unknown for the melt pool dynamics. In the above forward problem, the estimated or averaged values of these numbers are provided along with the training data. However, for complex multiphysics problems (e.g., LPBF and laser welding) it is very difficult to determine these equation parameters. The unique capability of the PINN algorithm is that it can learn these equation parameters if temperature and velocity data are provided for training at some time steps. This section focuses on addressing this problem through the data-driven discovery of partial differential equations (PDE) parameters.

This algorithm can learn the equation parameters of any PDEs of the form:

$$u_t + \mathcal{N}[u, \lambda] = 0 \qquad (22)$$

where $u(t, x)$ is the solution of the PDE, $\mathcal{N}[\cdot]$ is a non-linear differential operator and $\lambda$ denotes the equation parameters. Given the measurements of $\{t^i, x^i, y^i, u^i\}_{i=1}^{N}$ of the solution u, the equation parameter $\lambda$ can be inferred.

To test the validity of the algorithm, the effect of Argon flow on the melt pool dynamics problem is studied as a case study. The governing equations 1-4 are defined in the previous section. The objective of the study is to learn the values of governing equation parameters i.e Reynolds and Peclet numbers by using the measurement of the velocity field $\{t^i, x^i, y^i, u^i, v^i\}_{i=1}^{N}$.

The temperature and velocity data at 13,500 scattered collocation points per time step for 13 different time steps are provided to train the model (Figure 11). This data was generated from the COMSOL multiphysics software. Then, this data is used to train the fully connected PINN with 10 hidden layers and 50 neurons per hidden layer. The model was trained for 20 hours. The equation parameters ($R_{ey}$ and $P_{ec}$) can be learned by minimizing the mean squared error (MSE) loss of training data and residual of governing equations. The MSE loss of training data can be defined as:

$$\mathcal{L}_{Data} = \frac{1}{N}\sum_{n=1}^{N}\left(\left|u_{pred}^n(t^c,x^c,y^c) - u_{exact}^n\right|^2 + \left|v_{pred}^n(t^c,x^c,y^c) - v_{exact}^n\right|^2\right) \quad (23)$$

The residual loss for the governing PDEs used to train the network can be defined by Eq. 19.

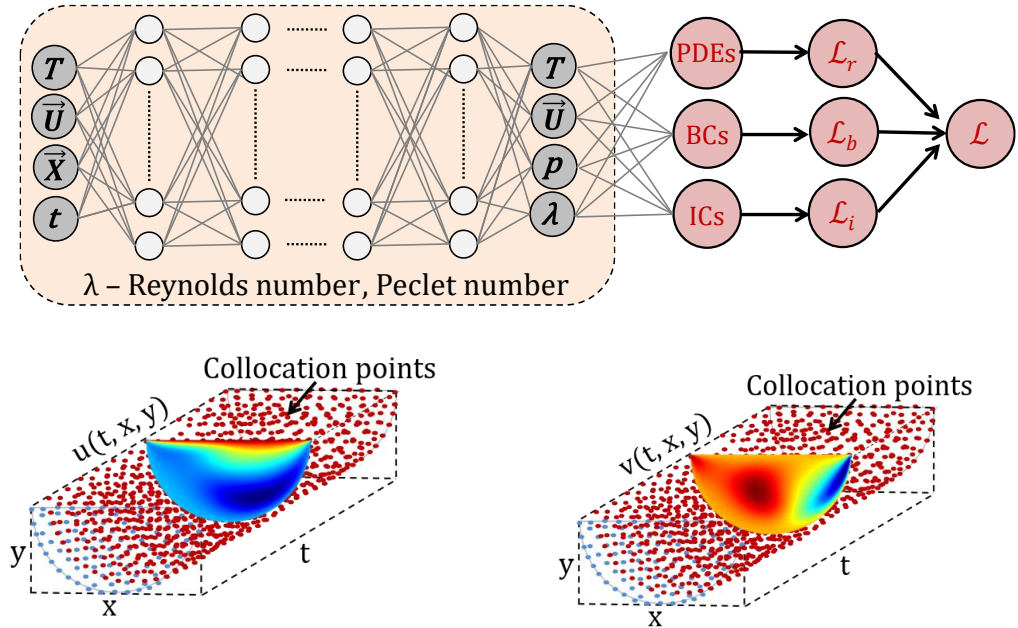

**Fig. 11:** The spatio-temporal training data-points for the u(t, x, y) and v(t, x, y) respectively.

Table 2 represents the predicted value and error percentage in the prediction of Rey and Pec. It is observed that the predicted values are in good agreement with the reference/estimated value. It should be noted that the inference of PINN is fast if the initial guess of equation parameters is closer to the actual value.

**Table 2**

The inferred Reynolds number (Rey) and Peclet number (Pec).

| Inferred model constants | Training time: 20 hrs. | | |
|---|---|---|---|
| | Reference | Inferred | Relative error (%) |
| Rey | 541 | 535.2 | 1.07 |
| Pec | 77 | 78.2 | 1.56 |

## 4. Discussions

The previous sections show the capability of PINNs to predict the melt pool dynamics including the velocity field without using any velocity training data. The PINN model is also capable of inferring governing equation parameters like Reynolds number and Peclet number if temperature and velocity data are used to train the model. For many practical multiphysics applications like LPBF, one can measure the temperature data by using IR cameras or pyrometers. The melt pool dynamics is very important for LPBF as it not only determines process stability but also affects the microstructures and the final material properties of the printed components. While physics-based simulation is the dominant approach to predict melt pool dynamics in literature, it has the inherent issue of very high computational cost due to the coupled energy and Navier-Stokes equations with the many source terms. In these scenarios, PINN provides a promising new approach to solving multiphysics problems which avoid the solving of highly non-linear PDE equations. Specific to LPBF, the following modeling steps can be followed to predict the melt pool dynamics (Figure 2):

- Perform a physics-based simulation with the known model constants of the governing equations for a shorter-time span to generate the training data of temperature and velocity for an inverse problem.
- The inverse problem may infer the model constants of the governing equations (e.g., the Reynolds number and Peclet number for the Navier-Stokes equations).
- With the inferred model constants of the governing equations and the temperature data (e.g., from high-speed IR cameras or pyrometers) the velocity and pressure field can be predicted for future longer-time steps. However, the magnitude of predicted pressure is shifted from the exact pressure by a constant value for this study. The difference can be attributed to the nature of the governing Navier-Stokes equation which contains only the derivative of the pressure due to which pressure prediction is always shifted from the exact value.
- With the calculated melt pool dynamics information, one may further predict other associated quantities like residual stress, microstructure, and properties of the printed materials in the future.

## 5. Conclusions

In this study, an innovative physics-informed machine learning approach by integrating neural networks with the governing physical laws to predict the melt pool dynamics such as temperature, velocity, and pressure is developed. The major advantages of this approach include the prediction of the velocity and pressure fields in the melt pool without using any priori training data of velocity and pressure, small training data, and short training time when compared to the conventional ML models, and its capability of inferring governing equation parameters. The key findings of this study can be enumerated as follows:

- PIML can be used to predict the accurate velocity field without having any priori training data of velocity for the Argon gas-driven melt pool dynamics in laser powder bed fusion.
- PIML can make a good qualitative prediction of pressure without any priori training data of pressure.
- The developed PIML model accurately predicts the temperature but struggles to accurately predict the velocity and pressure fields without utilizing the boundary data.

- It is observed that the PIML model can predict temperature, velocity, and pressure accurately by using only 850 data points and 7 time points, which is comparatively very data efficient as compared to the massive data required for a conventional pure data-driven ML model.
- The PIML model can be used to infer the governing equation parameters such as Reynolds and Peclet numbers by giving sufficient training data on velocity and temperature. It is observed that training time is proportional to the accuracy of the initial guess.

This study introduces a new approach to predicting the velocity and pressure field by using only the temperature data for training in LPBF. To check the capability of this algorithm, this study does not handle the complexity of LPBF processes. Future research may incorporate more complex physics, such as evaporation, natural convection, and Marangoni convection, to study their interactions in the melt pool.

**Acknowledgments**

The authors would like to thank the financial support of the National Science Foundation under the grant CMMI- 2152908.

**References**

[1] C. Panwisawas, C.L. Qiu, Y. Sovani, J.W. Brooks, M.M. Attallah, H.C. Basoalto, On the role of thermal fluid dynamics into the evolution of porosity during selective laser melting, Scripta Materialia 105 (2015) 14-17.
[2] D.D. Baere, M. Bayat, S. Mohanty, J. Hattel, Thermo-fluid-metallurgical modelling of the selective laser melting process chain, Procedia CIRP 74 (2018) 87-91.
[3] F.J. Gürtler, M. Karg, K.H. Leitz, M. Schmidt, Simulation of Laser Beam Melting of Steel Powders using the Three-Dimensional Volume of Fluid Method, Physics Procedia 41 (2013) 881-886.
[4] C.-C. Tseng, C.-J. Li, Numerical investigation of interfacial dynamics for the melt pool of Ti-6Al-4V powders under a selective laser, International Journal of Heat and Mass Transfer 134 (2019) 906-919.
[5] C. Panwisawas, C. Qiu, M.J. Anderson, Y. Sovani, R.P. Turner, M.M. Attallah, J.W. Brooks, H.C. Basoalto, Mesoscale modelling of selective laser melting: Thermal fluid dynamics and microstructural evolution, Computational Materials Science 126 (2017) 479-490.
[6] C.-J. Li, T.-W. Tsai, C.-C. Tseng, Numerical Simulation for Heat and Mass Transfer During Selective Laser Melting of Titanium alloys Powder, Physics Procedia 83 (2016) 1444-1449.
[7] H. Shen, P. Rometsch, X. Wu, A. Huang, Influence of Gas Flow Speed on Laser Plume Attenuation and Powder Bed Particle Pickup in Laser Powder Bed Fusion, JOM 72(3) (2020) 1039-1051.
[8] K. Kamimuki, T. Inoue, K. Yasuda, M. Muro, T. Nakabayashi, A. Matsunawa (2003) Behaviour of monitoring signals during detection of welding defects in YAG laser welding. Study of monitoring technology for YAG laser welding (Report 2), Welding International, 17:3, 203-210.

[9] R. Fabbro, S. Slimani, I. Doudet, F. Coste, F. Briand, Experimental study of the dynamical coupling between the induced vapour plume and the melt pool for Nd–Yag CW laser welding, Journal of Physics D: Applied Physics 39(2) (2006) 394.
[10] J. Greses, P.A. Hilton, C.Y. Barlow, W.M. Steen Plume attenuation under high power Nd:yttritium–aluminum–garnet laser welding. Journal of Laser Applications 1 February 2004; 16 (1): 9–15.
[11] J. Zou, W. Yang, S. Wu, Y. He, R. Xiao Effect of plume on weld penetration during high-power fiber laser welding. Journal of Laser Applications 1 May 2016; 28 (2): 022003.
[12] E. Rabinovich, H. Kalman, Generalized master curve for threshold superficial velocities in particle–fluid systems, Powder Technology 183(2) (2008) 304-313.
[13] A.B. Anwar, Q.-C. Pham, Selective laser melting of AlSi10Mg: Effects of scan direction, part placement and inert gas flow velocity on tensile strength, Journal of Materials Processing Technology 240 (2017) 388-396.
[14] B. Ferrar, L. Mullen, E. Jones, R. Stamp, C.J. Sutcliffe, Gas flow effects on selective laser melting (SLM) manufacturing performance, Journal of Materials Processing Technology 212(2) (2012) 355-364.
[15] A.B.S. Alquaity, B.S. Yilbas, Investigation of Spatter Trajectories in an SLM Build Chamber under Argon Gas Flow, Metals, 2022.
[16] S. Wang, Z. Su, L. Ying, X. Peng, S. Zhu, F. Liang, D. Feng, D. Liang, Accelerating magnetic resonance imaging via deep learning, IEEE.
[17] S. Min, B. Lee, S. Yoon, Deep learning in bioinformatics, Briefings in Bioinformatics 18(5) (2016) 851-869.
[18] T. Young, D. Hazarika, S. Poria, E. Cambria, Recent Trends in Deep Learning Based Natural Language Processing, arXiv pre-print server (2018).
[19] A. Torfi, Rouzbeh, Y. Keneshloo, N. Tavaf, Edward, Natural Language Processing Advancements By Deep Learning: A Survey, arXiv pre-print server (2021).
[20] A. Voulodimos, N. Doulamis, A. Doulamis, E. Protopapadakis, Deep Learning for Computer Vision: A Brief Review, Computational Intelligence and Neuroscience 2018 (2018) 7068349.
[21] A. Esteva, K. Chou, S. Yeung, N. Naik, A. Madani, A. Mottaghi, Y. Liu, E. Topol, J. Dean, R. Socher, Deep learning-enabled medical computer vision, Digital Medicine 4(1) (2021) 5.
[22] S.L. Brunton, B.R. Noack, P. Koumoutsakos, Machine Learning for Fluid Mechanics, Annual Review of Fluid Mechanics 52(1) (2020) 477-508.
[23] S.L. Brunton, Applying machine learning to study fluid mechanics, Acta Mechanica Sinica 37(12) (2021) 1718-1726.
[24] Z.Y. Wan, P. Vlachas, P. Koumoutsakos, T. Sapsis, Data-assisted reduced-order modeling of extreme events in complex dynamical systems, PLOS ONE 13(5) (2018) e0197704.
[25] K. Fukami, K. Fukagata, K. Taira, Assessment of supervised machine learning methods for fluid flows, Theoretical and Computational Fluid Dynamics 34(4) (2020) 497-519.
[26] L. Chen, X. Yao, C. Tan, W. He, J. Su, F. Weng, Y. Chew, N.P.H. Ng, S.K. Moon, In-situ crack and keyhole pore detection in laser directed energy deposition through acoustic signal and deep learning, Additive Manufacturing 69 (2023) 103547.
[27] S.M. Estalaki, C.S. Lough, R.G. Landers, E.C. Kinzel, T. Luo, Predicting defects in laser powder bed fusion using in-situ thermal imaging data and machine learning, Additive Manufacturing 58 (2022) 103008.

[28] A. Suzuki, Y. Shiba, H. Ibe, N. Takata, M. Kobashi, Machine-learning assisted optimization of process parameters for controlling the microstructure in a laser powder bed fused WC/Co cemented carbide, Additive Manufacturing 59 (2022) 103089.
[29] F. Ogoke, A.B. Farimani, Thermal control of laser powder bed fusion using deep reinforcement learning, Additive Manufacturing 46 (2021) 102033.
[30] J. Petrich, Z. Snow, D. Corbin, E.W. Reutzel, Multi-modal sensor fusion with machine learning for data-driven process monitoring for additive manufacturing, Additive Manufacturing 48 (2021) 102364.
[31] E. Westphal, H. Seitz, A machine learning method for defect detection and visualization in selective laser sintering based on convolutional neural networks, Additive Manufacturing 41 (2021) 101965.
[32] S. Liu, A.P. Stebner, B.B. Kappes, X. Zhang, Machine learning for knowledge transfer across multiple metals additive manufacturing printers, Additive Manufacturing 39 (2021) 101877.
[33] D.J. McGregor, M.V. Bimrose, C. Shao, S. Tawfick, W.P. King, Using machine learning to predict dimensions and qualify diverse part designs across multiple additive machines and materials, Additive Manufacturing 55 (2022) 102848.
[34] O.F. Ogoke, K. Johnson, M. Glinsky, C. Laursen, S. Kramer, A. Barati Farimani, Deep-learned generators of porosity distributions produced during metal Additive Manufacturing, Additive Manufacturing 60 (2022) 103250.
[35] S. Guo, M. Agarwal, C. Cooper, Q. Tian, R.X. Gao, W. Guo, Y.B. Guo, Machine learning for metal additive manufacturing: Towards a physics-informed data-driven paradigm, Journal of Manufacturing Systems 62 (2022) 145-163.
[36] R. Sharma, Y.B. Guo Computational Modeling and Physics-Informed Machine Learning of Metal Additive Manufacturing: State-of-the-Art and Future Perspective. Annual Review of Heat Transfer. 2021;24.
[37] M. Raissi, Deep Hidden Physics Models: Deep Learning of Nonlinear Partial Differential Equations, arXiv pre-print server (2018).
[38] E. Haghighat, M. Raissi, A. Moure, H. Gomez, R. Juanes, A physics-informed deep learning framework for inversion and surrogate modeling in solid mechanics, Computer Methods in Applied Mechanics and Engineering 379 (2021) 113741.
[39] B. Moseley, A. Markham, T. Nissen-Meyer, Finite Basis Physics-Informed Neural Networks (FBPINNs): a scalable domain decomposition approach for solving differential equations, arXiv pre-print server (2021).
[40] H. Gao, L. Sun, J.-X. Wang, PhyGeoNet: Physics-informed geometry-adaptive convolutional neural networks for solving parameterized steady-state PDEs on irregular domain, Journal of Computational Physics 428 (2021) 110079.
[41] Z. Fang, A High-Efficient Hybrid Physics-Informed Neural Networks Based on Convolutional Neural Network, IEEE Transactions on Neural Networks and Learning Systems 33(10) (2022) 5514-5526.
[42] R. Zhang, Y. Liu, H. Sun, Physics-informed multi-LSTM networks for metamodeling of nonlinear structures, Computer Methods in Applied Mechanics and Engineering 369 (2020) 113226.
[43] Y.A. Yucesan, F.A.C. Viana, Hybrid physics-informed neural networks for main bearing fatigue prognosis with visual grease inspection, Computers in Industry 125 (2021) 103386.

[44] L. Yang, X. Meng, G.E. Karniadakis, B-PINNs: Bayesian physics-informed neural networks for forward and inverse PDE problems with noisy data, Journal of Computational Physics 425 (2021) 109913.
[45] X. Jin, S. Cai, H. Li, G.E. Karniadakis, NSFnets (Navier-Stokes flow nets): Physics-informed neural networks for the incompressible Navier-Stokes equations, Journal of Computational Physics 426 (2021) 109951.
[46] C.J. Arthurs, A.P. King, Active training of physics-informed neural networks to aggregate and interpolate parametric solutions to the Navier-Stokes equations, Journal of Computational Physics 438 (2021) 110364.
[47] S. Cuomo, Vincenzo, F. Giampaolo, G. Rozza, M. Raissi, F. Piccialli, Scientific Machine Learning through Physics-Informed Neural Networks: Where we are and What's next, arXiv pre-print server (2022).
[48] Q. Zhu, Z. Liu, J. Yan, Machine learning for metal additive manufacturing: predicting temperature and melt pool fluid dynamics using physics-informed neural networks, Computational Mechanics 67(2) (2021) 619-635.
[49] S. Liao, T. Xue, J. Jeong, S. Webster, K. Ehmann, J. Cao, Hybrid thermal modeling of additive manufacturing processes using physics-informed neural networks for temperature prediction and parameter identification, Computational Mechanics (2023).
[50] A. Aggarwal, S. Patel, A. Kumar, Selective Laser Melting of 316L Stainless Steel: Physics of Melting Mode Transition and Its Influence on Microstructural and Mechanical Behavior, JOM 71(3) (2019) 1105-1116.
[51] C. Tang, K.Q. Le, C.H. Wong, Physics of humping formation in laser powder bed fusion, International Journal of Heat and Mass Transfer 149 (2020) 119172.
[52] M. Raissi, A. Yazdani, G.E. Karniadakis, Hidden fluid mechanics: Learning velocity and pressure fields from flow visualizations, Science 367(6481) (2020) 1026-1030.
[53] M. Raissi, P. Perdikaris, G.E. Karniadakis, Physics-informed neural networks: A deep learning framework for solving forward and inverse problems involving nonlinear partial differential equations, Journal of Computational Physics 378 (2019) 686-707.